%% file: egpaper_final.tex
\documentclass[10pt,twocolumn]{article}
\usepackage{3dv}
\usepackage{times}
\usepackage{epsfig}
\usepackage{graphicx}
\usepackage{amsmath}
\usepackage{amssymb}

\usepackage[numbers,sort&compress]{natbib}
\usepackage{color}
\usepackage{wrapfig}
\usepackage{booktabs}
\usepackage{tabularx}
\usepackage{float}
\usepackage{adjustbox}
\usepackage{comment}
\usepackage{pdfpages}

\usepackage[pagebackref=true,breaklinks=true,letterpaper=true,colorlinks,bookmarks=false]{hyperref}

\newcommand{\VS}[1]{\textcolor{red}{#1}}

\newcommand{\srn}{\textsc{srn}}
\newcommand{\seg}{\textsc{seg}}
\newcommand{\hn}{\textsc{hn}}

\threedvfinalcopy % *** Uncomment this line for the final submission

\def\threedvPaperID{****} % *** Enter the 3DV Paper ID here

\ifthreedvfinal\fi
\begin{document}

%%%%%%%%% TITLE
\title{Semantic Implicit Neural Scene Representations With Semi-Supervised Training}

\author{Amit Pal Singh Kohli*\\
Stanford University, UC Berkeley\\
{\tt\small apkohli@berkeley.edu}
\and
Vincent Sitzmann*\\
Stanford University, MIT CSAIL\\
{\tt\small sitzmann@cs.stanford.edu}
\and
Gordon Wetzstein\\
Stanford University\\
{\tt\small gordon.wetzstein@stanford.edu}
}

\maketitle
\thispagestyle{empty}

%%%%%%%%% ABSTRACT
\begin{abstract}
The recent success of implicit neural scene representations has presented a viable new method for how we capture and store 3D scenes. Unlike conventional 3D representations, such as point clouds, which explicitly store scene properties in discrete, localized units, these implicit representations encode a scene in the weights of a neural network which can be queried at any coordinate to produce these same scene properties.
Thus far, implicit representations have primarily been optimized to estimate only the appearance and/or 3D geometry information in a scene. We take the next step and demonstrate that an existing implicit representation (SRNs) ~\cite{sitzmann_scene_2019} is actually multi-modal; it can be further leveraged to perform per-point semantic segmentation while retaining its ability to represent appearance and geometry.
To achieve this multi-modal behavior, we utilize a semi-supervised learning strategy atop the existing pre-trained scene representation. Our method is simple, general, and only requires a few tens of labeled 2D segmentation masks in order to achieve dense 3D semantic segmentation.
We explore two novel applications for this semantically aware implicit neural scene representation: 3D novel view and semantic label synthesis given only a single input RGB image or 2D label mask, as well as 3D interpolation of appearance and semantics.

\begin{comment}	
Biological vision infers multi-modal 3D representations that support reasoning about scene properties such as materials, appearance, affordance, and semantics in 3D.
%
These rich representations enable us humans, for example, to acquire new skills---such as the learning of a new semantic class---with extremely limited supervision.
%	
Motivated by this ability of biological vision, we demonstrate that 3D-structure-aware representation learning leads to multi-modal representations that enable 3D semantic segmentation with extremely limited, 2D-only supervision.
%
Building on emerging implicit neural scene representations, which have been developed for modeling the shape and appearance of 3D scenes supervised exclusively by posed 2D images, we are first to demonstrate a representation that jointly encodes shape, appearance, and semantics in a 3D-structure-aware manner. 
%
Surprisingly, we find that only a few tens of labeled 2D segmentation masks are required to achieve dense 3D semantic segmentation using a semi-supervised learning strategy.
%
We explore two novel applications for our semantically aware implicit neural scene representation: 3D novel view and semantic label synthesis given only a single input RGB image or 2D label mask, as well as 3D interpolation of appearance and semantics.
\end{comment}
\end{abstract}

%%%%%%%%% BODY TEXT
\section{Introduction}
\label{sec:introduction}
\input{sections/introduction}

\begin{figure*}[t]
	\centering
	\includegraphics[width=\textwidth]{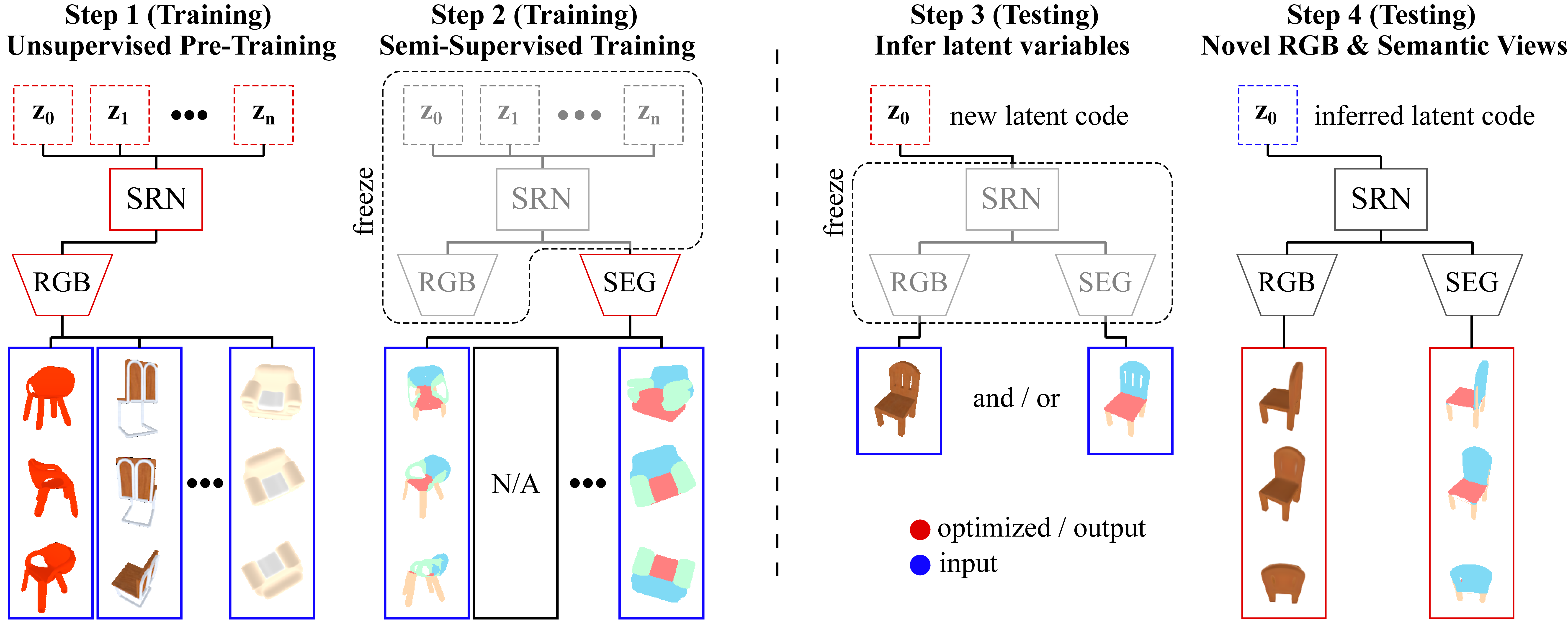}
	\caption{Overview of the proposed semi-supervised method applied to SRNs. From left to right: (1) We pre-train the SRN for novel view synthesis of RGB images using a large dataset of 2D posed images in an autodecoder-framework~\cite{ParkFSNL2019}, where each object instance is represented by its own code vector $\mathbf{z}_i$. (2) We then freeze code vectors and weights of the SRN and train a linear segmentation classifier on the features, using 30 human-annotated semantic labels. (3) At test time, given a \emph{single} posed RGB image and/or label mask of an instance unseen  at training time, we infer the latent code of the novel object. (4) Subsequently, we may render multi-view consistent novel RGB and semantic segmentation views of the object instance.}
	\label{fig:pipeline_overview}
\end{figure*}

\section{Related Work}
\label{sec:related_work}
\input{sections/related_work}
\section{Method}
\label{sec:method}
\input{sections/method}

\input{sections/analysis}

\section{Discussion}
\label{sec:discussion}
\input{sections/discussion}

{\small
\bibliographystyle{ieee}
\bibliography{SSRNsBib1}
}


\section*{Supplementary material}

# Semantic Implicit Neural Scene Representations with Semi-supervised Training

Amit Pal Singh Kohli
Stanford University, UC Berkeley
apkohli@berkeley.edu

Vincent Sitzmann
Stanford University, MIT CSAIL
sitzmann@cs.stanford.edu

Gordon Wetzstein
Stanford University
gordon.wetzstein@stanford.edu

## 1. Model Implementation Details

Here we specify the architectures and training details for both the SRN-based and U-Net based models as well as the baseline model used in the main paper. We implement all models in PyTorch. We train SRN-based models on Nvidia RTX8000 GPUs, and other models on Pascal TitanX GPUs.

**SRN-based models.** The basis for our architecture comes from the Scene Representation Network as proposed by Sitzmann et al. [6]. The SRN as well as the *RGB Renderer* are implemented as 4-layer MLPs with 256 units each, ReLU nonlinearities, and LayerNorm before each nonlinearity. The raymarcher is implemented as an LSTM [2] with 256 units. We ray march for 10 steps. We train our models using ADAM with a learning rate of $4e-4$. For our proposed 3D representation learning method (SRN+Linear), the key insight is that we take the pre-trained features $v \in \mathbb{R}^{256}$ from the neural scene representation and use a simple linear transformation to map those features to class probabilities for each pixel. For an object with $c$ semantic classes, the optimization parameters are matrix $W \in \mathbb{R}^{256 \mathrm{x} c}$ and bias $\mathrm{b} \in \mathbb{R}^c$. Specifically, in the case of chairs $c = 6$ and for tables $c = 11$. SRN-based models are trained for 20k steps at a resolution of 64 with a batch size of 92, and trained for another 85k steps at a resolution of 128 with a batch size of 16. Image reconstruction loss and cross-entropy loss are weighted $200 : 8$, such that their magnitudes are approximately equal.

**U-Net-based models.** We use a classic and effective approach for semantic segmentation, a UNet [5]. Specifically, we utilize an architecture based on the one presented in Isola et al. [3], which is shown in 3. Each downsampling layer consists of one stride-one convolutional layer, followed by one stride-two convolutional layer. Each upsampling layer consists of one stride-two transpose convolutional layer, followed by one stride-one convolutional layer. We use BatchNorm and LeakyReLU activations after each convolutional block and dropout with a rate of $0.1$. We train this model using the Adam optimizer with a learning rate of $4e-4$ and a batch size of 64 until convergence of validation error after about 80k iterations or 20 epochs.

**Baseline model.** For this baseline we implement the model exactly as specified by Tatarchenko et al. [7]. Implementation information can be found on their github: https://github.com/lmb-freiburg/mv3d.

## 2. Additional U-Net Baseline

Here we introduce an additional baseline in order to address the naïve approach of training a 2D to 2D segmentation model on the output of an SRN. This approach has the same input-output capability (single view in, arbitrary appearance and semantic views out) as our proposed model, but does not create a semantically-informed 3D representation and instead infers semantics after rendering images from an existing 3D representation. We demonstrate that the joint representation used by our model allows it to outperfom the baseline in a low data regime. In this regime, the baseline overfits very quickly and performs poorly on the test set. Furthermore, because it lacks the 3D strucucture that is baked into the representation from our model, the baseline tends to fail in classifying difficult views in which key features of the object are occluded.

In 1 we run an experiment training the baseline given increasing amounts semantically labeled data. For each instance in the variable sized datasets, there are 3 randomly sampled views per each chair. The models are trained identically with early stopping based on a validation set during training. Each model is then evaluated on the mIOU metric. Clearly the baseline's ability to perform segmentation is heavily dependent on the amount of data it has to train on. The baseline only matches the performance of our model when it has access to more than $20\%$ of all training instances, whereas our model only requires

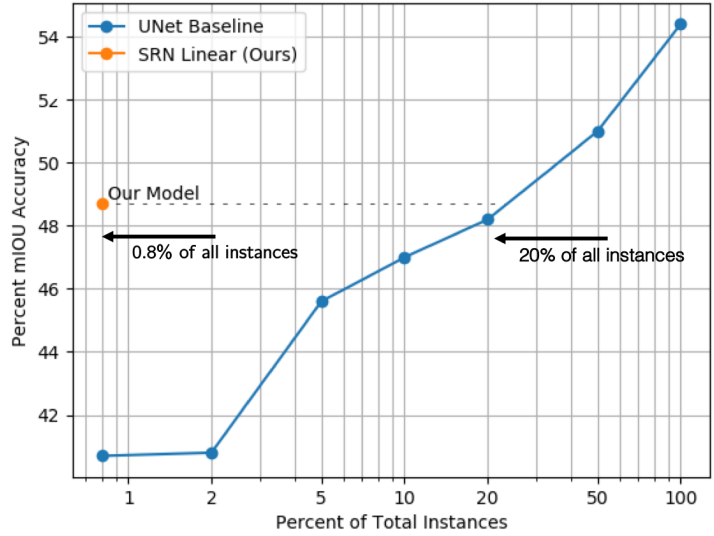

Figure 1. Graph demonstrating the UNet's reliance on data for segmentation performance. X-axis is percentage of total instances of objects in the training set, where each instance has 3 random, posed views. Y-axis is the mIOU measure averaged over all test instances. This experiment was performed for the chair object class with a total of 1214 instances.

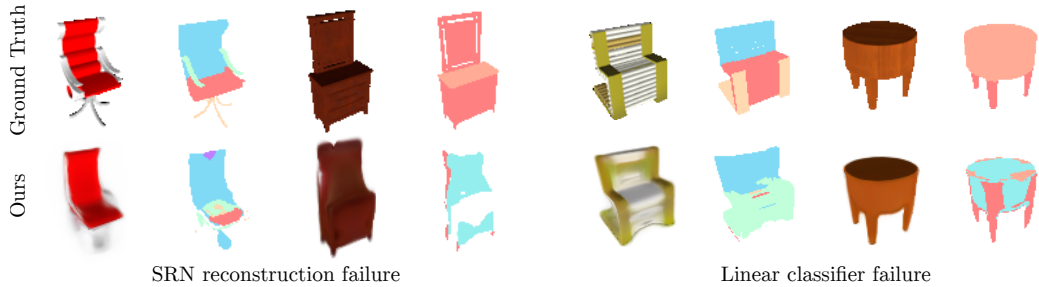

Figure 2. Failure cases.

0.8% of all training instances. Qualitative comparisons further emphasize this result and can be found in the attached video. For this baseline, we use the same U-Net architecture as described above in 1.

## 3. Failure cases.

Fig. 2 displays failure cases of the proposed approach. The proposed approach inherits limitations and failure cases of scene representation networks, such as failure to reconstruct strong out-of-distribution samples or objects with small gaps or high-frequency geometric detail. In these cases, the semantic segmentation fails as well. In the semi-supervised regime, the linear classifier sometimes fails to assign the correct class even if geometry and appearance were reconstructed correctly, which we attribute to its limited representative power. We note that as both appearance-based 3D neural scene representation methods as well as semi-supervised representation learning methods further develop, these failure cases will improve. Additional failure case examples can be found in the supplemental video.

## 4. Rendering

For our dataset we use Partnet [4] and Shapenet [1]. For each instance we align a Partnet and Shapenet model and render them using Blender; the Shapenet instance is used for the RGB views and the Partnet instance is used for the corresponding segmentation masks. All camera matrices were also written out in this process. The train-val-test split is from the semantic segmentation task laid out in Mo et al. [4].

## 5. Baseline Architecture Pix2Pix (Isola et al. [3])

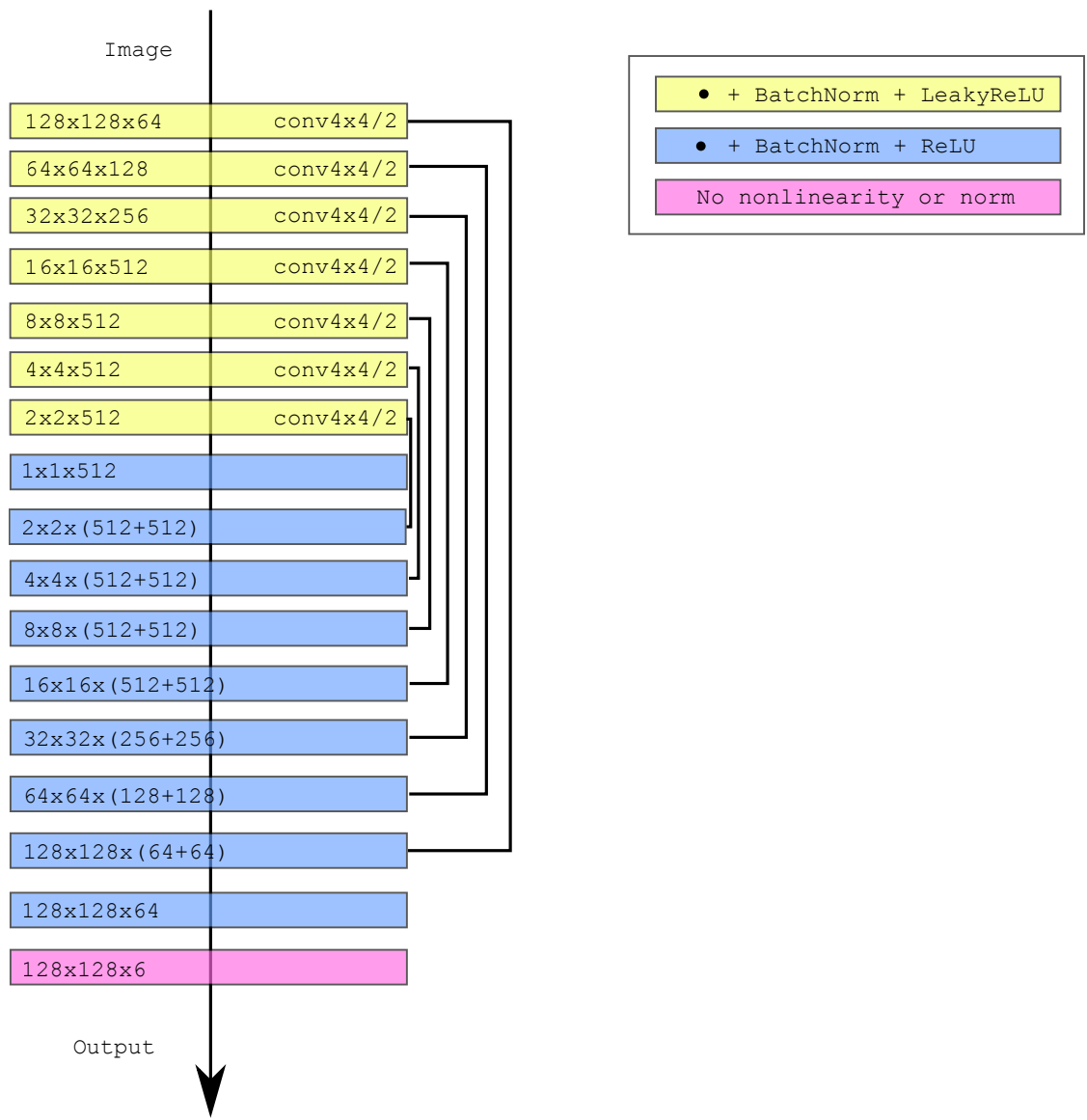

Figure 3. Architectural details of the image-to-image translation baseline model based on Pix2Pix by Isola et al. [3].



\end{document}

%% file: sections/introduction.tex
\begin{figure}[t]
	\centering
	\includegraphics[width=\linewidth]{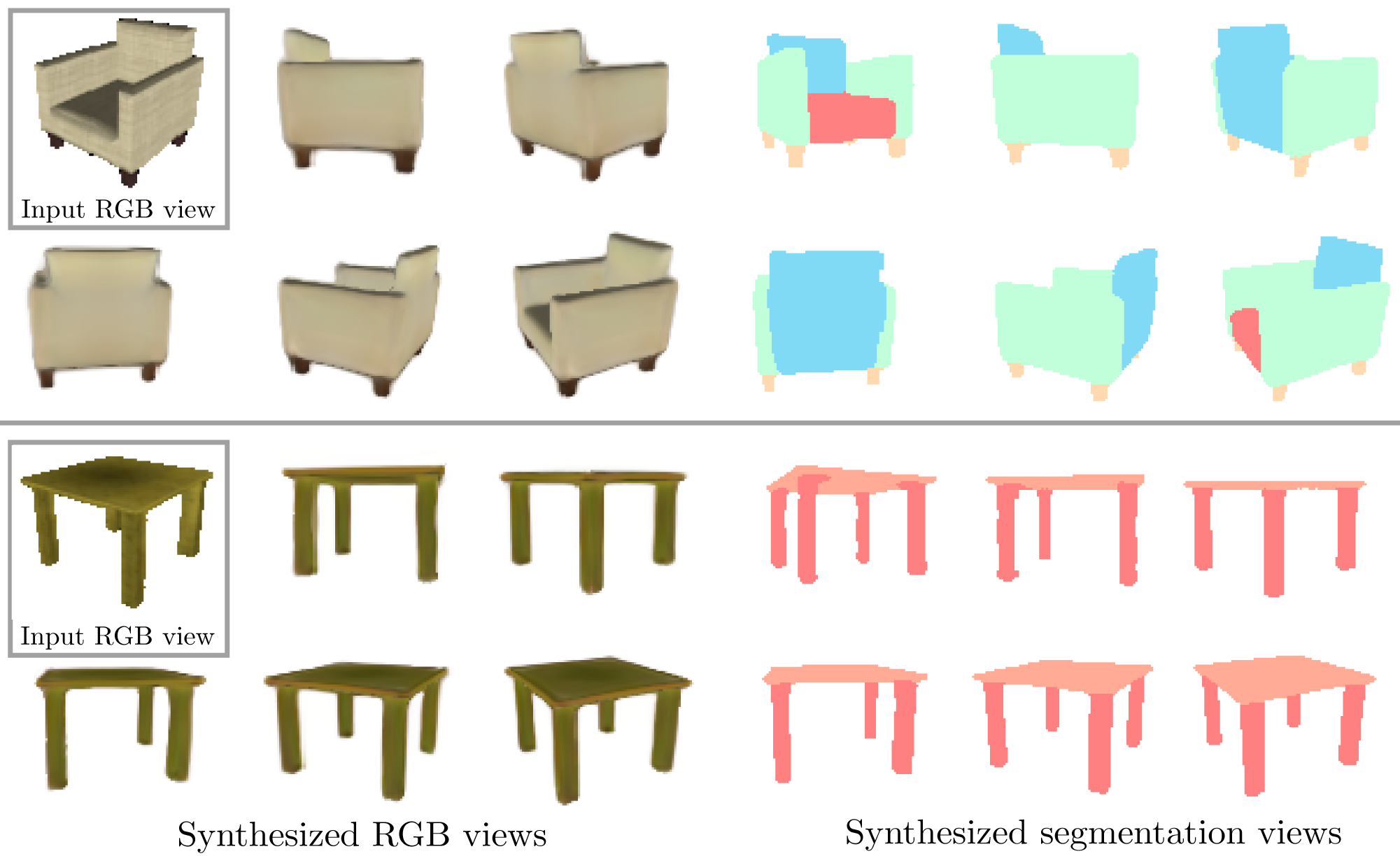}
	\caption{A single RGB image of an unseen object (upper left) is fed into the network, which is then capable of synthesizing perspectively consistent 3D RGB views (left) and semantic segmentation labels (right) of the object.}
	\label{fig:teaser}
\end{figure}

Humans have the innate ability to recall numerous properties of a scene, such as materials, color, geometry, lighting, or semantic classes from just a single observation. Furthermore, we are able to learn new attributes  quickly, with little supervision; a person does not need to be told the name of an object thousands of times in order to recognize it.
Thus, we say that humans have a \emph{multi-modal}  representation of the world. Specifically, a representation is multi-modal if it provides information about several different modalities without significant additional computation. 
For instance, a 3D scene representation which provides appearance and semantic class at every 3D coordinate is multi-modal. This is in contrast to a representation which stores appearance only: while semantic information could be estimated from appearance, this would require significant additional computation.

A similar ability for multi-modal vision and learning from limited supervision is also crucial for many tasks in computer vision, robotics, and autonomous driving. 
In these applications, algorithms must reason about a 3D scene given only partial information, such as a single image.
In robotic grasping, for instance, a robot has to simultaneously reason about the 3D geometry, appearance, and semantic structure of an object in order to choose the optimal grasping point.
Human labeling is expensive, and these applications would thus greatly benefit from label-efficient learning approaches.

Recent progress in representation learning has enabled competitive performance on 2D tasks when only a limited amount of training data is available~\cite{zhu2009introduction,Chapelle2009,henaff2019data,bachman2019learning,donahue2019large}. 
Here, 2D feature extractors are trained with massive amounts of unlabeled data on a surrogate task. Once the representation is learned, a limited amount of training data can be sufficient to train a simple classifier on the pre-trained feature representation~\cite{henaff2019data}.
While these approaches are applicable to 2D image-based problems, they do not build a 3D-structure-aware representation. 
Given a single image observation, they are incapable of making predictions about unseen perspectives of the scene or occluded parts, a task that is critical to 3D scene understanding and interaction.

Concurrently, 3D implicit neural scene representations are an emerging paradigm to tackle problems in inverse graphics and 3D computer vision~\cite{TatarchenkoDB16,DosovSTB2017,eslami2018neural,yao_3d-aware_2018,ZhouTFFS2018,lsiTulsiani18,KimGTXTNPRZT2018,meshry_neural_nodate,SriniTBRNS2019,SitzmTHNWZ2019,FlynnBDDFOST2019,sitzmann_scene_2019,Phuoc2019GAN,LombaSSSLS2019,ParkFSNL2019,niemeyer2019occupancy}.  
Given 2D image observations, these approaches aim to infer a 3D-structure-aware representation of the underlying scene that enables prior-based predictions about occluded parts.
These scene representations have thus far been primarily explored for applications in view synthesis, but not for scene understanding.
A na\"ive approach would be to generate arbitrary perspectives of a scene from limited observations and then apply standard 2D methods for semantic segmentation or other tasks.
Such image-based approaches, however, fail to infer a compact, multi-modal representation that would allow for joint reasoning about \emph{all} aspects of the scene.

Here we view the recently proposed scene representation networks (SRNs) from a representation learning perspective in order to infer multi-modal, compact 3D representations of objects from 2D images.
We take the latent 3D feature representation of SRNs, learned in an unsupervised manner given only posed 2D RGB images, and map them to a set of labeled semantic segmentation maps.
We find that for a simple mapping, we are able to achieve dense 3D semantic segmentation given just a few tens of these semantic segmentation labels.
This unique combination of unsupervised, 3D-structure-aware pre-training and supervised fine-tuning enables multi-view consistent view synthesis and semantic segmentation (see Fig.~\ref{fig:teaser}). 
Our approach further enables several other novel applications, including interpolation of 3D segmentation labels as well as 3D view and semantic label synthesis from just a single observed image or semantic mask.
To summarize, we make the following key contributions:

\begin{itemize}	
	\item We develop a method for learning a semantically and 3D-structure-aware neural scene representation.
	
	\item In a semi-supervised learning framework, we demonstrate that the resulting representation can be leveraged to perform dense 3D semantic segmentation from only 2D observations, given as few as 30 semantic segmentation masks. We demonstrate that features learned by the 3D neural scene representation far outperform a neural scene representation without 3D structure.
	
	\item We demonstrate both multi-view consistent renderings and 3D point clouds of semantic segmentation masks, including parts of the object that are occluded in the observation. 
	
	\item We perform joint interpolation of geometry, appearance, and semantic labels, and demonstrate how a neural scene representation can be inferred from either a color image or a semantic segmentation mask.
\end{itemize}

%% file: sections/related_work.tex
Inferring properties of 3D environments given limited amounts of labeled training data has been a long-standing challenge in the computer vision community. 
Our approach takes a step towards this goal by combining insights from representation learning, neural scene representations, and 3D computer vision.
Each of these fields builds on extensive literature, which we summarize as follows.

\paragraph{3D Computer Vision.} Deep-learning-based models for geometry reconstruction were among the first to propose 3D-structured latent spaces to enable 3D reasoning about scenes.
Discretization-based techniques use voxel grids
\cite{kar2017learning,tulsiani2017multi,3dgan,GadelhaMW17,qi2016volumetric,pix3d,rezende2016unsupervised,choy20163d}, octree hierarchies \cite{Riegler2017OctNet,TatarchenkoDB17,Haene2019}, point clouds \cite{qi2017pointnet,pmlr-v80-achlioptas18a,TatarchenkoDB16}, multiplane images \cite{ZhouTFFS2018}, patches \cite{abs-1802-05384}, or meshes \cite{Jack2018,rezende2016unsupervised,kato2018neural,Kanazawa18categorymesh}.
Methods based on function spaces continuously represent space as the decision boundary of a learned binary classifier \cite{mescheder2018occupancy} or a continuous signed distance field \cite{ParkFSNL2019,genova2019learning,deng2019cvxnets}.
While these methods model the underlying 3D geometry of a scene, they do not model aspects of the scene other than geometry.

\paragraph{2D Representation Learning.} A large body of work explores self-supervised representation learning on images~\cite{Chapelle2009,donahue2019large,henaff2019data,bachman2019learning,jing2019self,kingma2014semi,rasmus2015semi,dai2017good,doersch2015unsupervised,zhang2016colorful,noroozi2016unsupervised,larsson2017colorization,wu2018unsupervised,donahue2016adversarial,radford2015unsupervised,oord2018representation}. 
These approaches have yielded impressive results on 2D tasks such as bounding box detection, 2D image segmentation, and image classification.
However, none of these approaches builds a 3D-structure-aware representation.
This lack of 3D inductive bias makes these approaches incapable of reasoning about multi-view consistency or object parts occluded in the input image.
Fundamentally, 2D representation learning is therefore incapable of supporting 3D semantic labeling from 2D input.

\paragraph{Neural Scene Representations.} 
A recent line of work reconstructs both appearance and geometry given only 2D images and their extrinsic and intrinsic camera parameters. 
Auto-encoder-like methods only weakly model the latent 3D structure of a scene~\cite{TatarchenkoDB16,worrall2017interpretable}.
Generative Query Networks~\cite{eslami2018neural,kumar2018consistent} introduce a probabilistic reasoning framework that models uncertainty due to incomplete observations, but both the scene representation and the renderer are oblivious to the scene's 3D structure.
Some recent work explores voxel grids as a scene representation~\cite{SitzmTHNWZ2019,tung2018learning,phuoc2018rendernet,zhu2018visual,Phuoc2019GAN}. 
Our approach builds on recent continuous, 3D-structure-aware scene representations~\cite{,thies2019neural,ParkFSNL2019,sitzmann_scene_2019, sitzmann2020implicit,genova2019learning,genova2019deep, atzmon2020sal, michalkiewicz2019implicit,gropp2020implicit,jiang2020local,peng2020convolutional,chabra2020deep,mescheder2018occupancy,saito2019pifu,mildenhall2020nerf,chen2019learning,oechsle2019texture,niemeyer2020differentiable,sitzmann2020metasdf}. For an in-depth review of neural scene representations, see ~\cite{tewari2020state}. 
BAE-Net \cite{chen2019bae} learns to perform 3D semantic segmentation in an unsupervised manner, but requires ground-truth 3D information at training time as well as 3D input at test time. Further, the proposed architecture has a specific inductive bias for learning semantic segmentation from occupancy prediction, and does not generalize to other modalities, such as appearance.

\paragraph{Semantic Segmentation.} 
The advent of deep learning has had a transformative impact on the field of semantic segmentation.
Seminal work by Long et al.~\cite{long2015fully} introduced fully convolutional neural networks for pixel-level semantic labeling. 
Numerous CNN-based approaches further refined this initial idea~\cite{chen2014semantic,zheng2015conditional,yu2015multi,ronneberger2015u}.
Recent work in this area has increasingly incorporated ideas from 3D computer vision. Semantic segmentation has thus been formulated in cases where both geometry and color information are available~\cite{dai20183dmv,valentin2015semanticpaint,vineet2015incremental, yi_gspn:_2018}.
However, these methods operate on point clouds or voxel grids and therefore rely on explicit geometry representations.
To the best of our knowledge, no semantic segmentation approach infers 3D semantic labels given a 2D RGB image, which our method enables.

%% file: sections/method.tex
Here we develop a semantically-aware implicit neural scene representation by leveraging an existing pre-trained scene representation  with a small set of semantically labeled data. For our experiments in Sec. \ref{sec:analysis}, we specify Scene Representation Networks (SRNs)~\cite{sitzmann_scene_2019} as the backbone scene representation. However, we make clear that our method can apply to any feature-based neural scene representation.

\begin{comment}
, we first review scene representation networks (SRNs)~\cite{sitzmann_scene_2019} and then demonstrate how we can extend SRNs to perform 3D semantic segmentation.
%
Finally, we view SRNs through the lense of representation learning and apply a semi-supervised learning strategy. This yields 3D semantic segmentation from 2D RGB observations and their camera parameters, given an extremely limited number of semantic segmentation masks.
\end{comment}

%%%%%%%%%%%%%%%%%%%%%%%%%%%%%%%%%%%%%%%%%%%%%%%%%%%%%%%%%%%%%%%%%%%%%%%%%%%%%%%%%%%%%%%%%%%%%%%%
\subsection{Implicit Neural Scene Representations}
Our method begins with pre-training an existing implicit neural scene representation. In general, we only require that it contain a feature representation, $\mathbf{v}$, for each point of interest in 3D space. In cases where intermediate features are not explicitly considered, such as in ~\cite{ParkFSNL2019} or ~\cite{TatarchenkoDB16}, we can extract these features as an intermediate layer of the network architecture. Here, we choose to use SRNs and provide a short review in order to make our method more clear.

\paragraph{Scene Representation Networks}
Scene Representation Networks are a continuous, 3D-structure aware neural scene representation.
They enable reconstruction of 3D appearance and geometry, trained end-to-end from only 2D images and their camera poses, without access to depth or shape.
The key idea of SRNs is to encode a scene in the weights $\mathbf{w} \in \mathbb{R}^l$ of a fully connected neural network, the \srn{} itself.
To this end, a scene is modeled as a function that maps world coordinates $\mathbf{x}$ to a feature representation of local scene properties $\mathbf{v}$:
\begin{align}
\srn: \mathbb{R}^3 \to \mathbb{R}^n, \quad \mathbf{x} \mapsto \srn(\mathbf{x}) = \mathbf{v}.
\label{eq:srn}
\end{align}
Images are synthesized from this 3D representation via a differentiable neural renderer consisting of two parts.
The first is a differentiable ray marcher which finds intersections of camera rays with scene geometry by marching along a ray away from a camera. At every step, it queries \srn{} at the current world coordinates and translates the resulting feature vector into a step length.
Finally, \srn{} is queried a final time at the regressed ray intersection points, and 
the resulting feature vector $\mathbf{v}$ is mapped to an RGB color via a fully connected neural network, which we refer to as the \emph{RGB Renderer}.
Due to the differentiable rendering, SRNs may be trained given only 2D camera images as well as their intrinsic and extrinsic camera parameters. 

To generalize across a class of objects, it is assumed that the weights $\mathbf{w}_j$ of \srn{}s that represent object instances within the same class lie in a low-dimensional subspace of $\mathbb{R}^l$, permitting us to represent each object via an embedding vector $\mathbf{z}_j \in \mathbb{R}^k$, $k<l$. 
A hypernetwork~\cite{ha2016hypernetworks} \hn{} maps embedding vectors $\mathbf{z}_j$ to the weights $\mathbf{w}_j$ of the respective scene representation network:
\begin{align}
\hn: \mathbb{R}^k \to \mathbb{R}^l, \quad \mathbf{z}_j \mapsto \hn (\mathbf{z}_j)=\mathbf{w}_j.
\end{align}
\hn{} thus learns a prior over the weights of scene representation networks and thereby over scene properties.
To infer the scene representation of a new scene or object, an embedding vector $\mathbf{z}$ is randomly initialized, the weights of \hn{} and the differentiable rendering are frozen, and $\mathbf{z}$ is optimized to obtain a new scene embedding via minimizing image reconstruction error.

%%%%%%%%%%%%%%%%%%%%%%%%%%%%%%%%%%%%%%%%%%%%%%%%%%%%%%%%%%%%%%%%%%%%%%%%%%%%%%%%%%%%%%%%%%%%%%%%
%\subsection{SRNs for Semantic Segmentation}
\subsection{Semantically-aware Scene Representations}
\label{subsec:seg}
We formalize dense 3D semantic segmentation as a function that maps a world coordinate $\mathbf{x}$ to a distribution over semantic labels $\mathbf{y}$.
This can be seen as a generalization of point cloud- and voxel-grid-based semantic segmentation approaches~\cite{dai20183dmv,qi2017pointnet,qi_pointnet++:_2017}, which label a discrete set of world coordinates, sparsely sampling an underlying, continuous function.
To leverage our pre-trained scene representation for semantic segmentation, we define the \emph{Segmentation Renderer} \seg{}, a function that maps a feature vector $\mathbf{v}$ to a distribution over class labels $\mathbf{y}$:
\begin{align}
\seg: \mathbb{R}^n \to \mathbb{R}^m, \quad \mathbf{v} \mapsto \seg (\mathbf{v})=\mathbf{y}.
\end{align}
For SRNs, this amounts to adding a \emph{Segmentation Renderer} in parallel to the existing \emph{RGB Renderer}.
Since $\mathbf{v}$ is a function of $\mathbf{x}$, we may enforce a per-pixel cross-entropy loss on the \seg{} output at any world coordinate $\mathbf{x}$:
\begin{equation}
\mathcal{L}_{\text{co}} = \sum_{j=1}^{c}\hat{y}_{j}\log{\sigma(\seg(\mathbf{v})})
\end{equation}
where $\hat{y}_{j}$ is a one-hot ground-truth class label with $c$ number of classes, and $\sigma$ is the softmax function.
We can now  train the segmentation renderer end-to-end composed with the same architecture used to pre-train the scene representation. When we apply this to SRNs, the features $\mathbf{v}$ are supervised to carry semantic information via the \emph{Segmentation Renderer}, in addition to the existing geometry information via the ray-marcher and RGB information via the \emph{RGB Renderer}.
At test time, this formulation with SRNs allows us to infer a code vector $\mathbf{z}$ from either RGB information, semantic segmentation information, or both. In any of these cases, a new code vector is inferred by freezing all network weights, initializing a new code vector $\mathbf{z}$, and optimizing $\mathbf{z}$ to minimize image reconstruction and/or cross entropy losses, see Fig.~\ref{fig:pipeline_overview}, Step 3. 

%%%%%%%%%%%%%%%%%%%%%%%%%%%%%%%%%%%%%%%%%%%%%%%%%%%%%%%%%%%%%%%%%%%%%%%%%%%%%%%%%%%%%%%%%%%%%%%%

%\subsection{Semi-Supervised Learning with SRNs}
\subsection{Semi-Supervised Learning of Semantically-aware Scene Representations}
\label{subsec:semisup}
While training end-to-end with a segmentation renderer on a large dataset of human-labeled images is straightforward, it has a significant weakness: it relies on a massive amount of labeled semantic data.
Such labeled data may be difficult to obtain for a variety of different computer vision tasks.
Moreover, it is desirable for an independent agent to infer an understanding of the different modes of an object it has not encountered. Such an unsupervised exploration cannot rely on thousands or millions of interactions with each object class to infer semantic properties.

\begin{figure}
	\centering
	\includegraphics[height=0.65\linewidth]{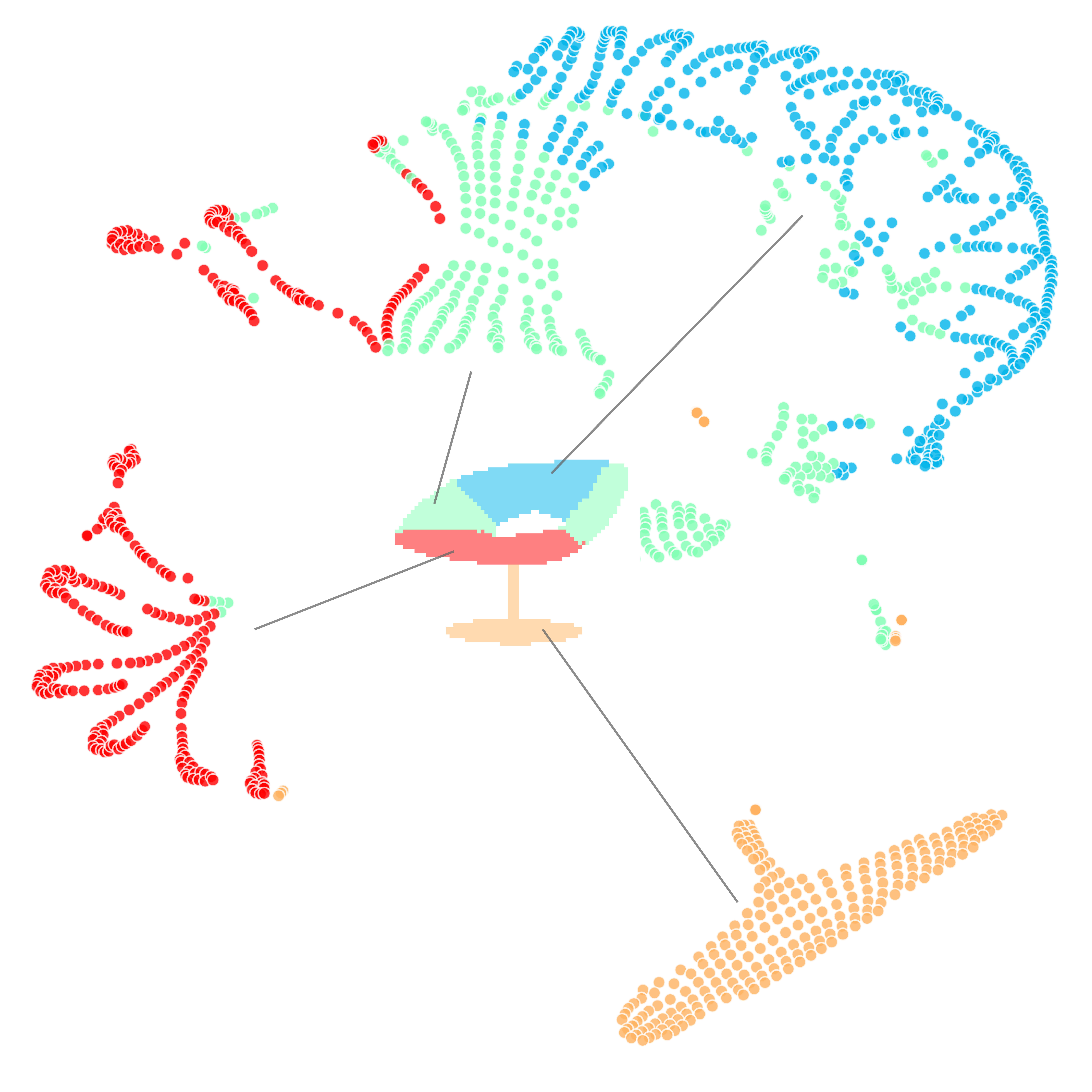}
	
	\caption{A t-SNE plot of features from a pre-trained SRN along with the test set chair that it represents. Each feature is labeled by a semantic part indicated on the chair. There is a clear clustering of the transformed features based on their corresponding class.}
	\label{fig:tsne}
\end{figure}

Inspired by 2D representation learning approaches ~\cite{Chapelle2009,henaff2019data,bachman2019learning,donahue2019large}, we view the original task of the implicit neural scene representation as enforcing features to encode information about appearance and geometry, and hypothesize that these same features will also be useful for the downstream task of dense 3D semantic segmentation. 
To further support this, as well as motivate our choice of SRNs, we plot the t-SNE ~\cite{maaten2008visualizing} embeddings of the features $\mathbf{v}$ of a pretrained SRN. We color each embedding according to the semantic class of the point in the scene it represents. In Fig. \ref{fig:tsne} we see that features of the same semantic class are naturally clustered, which suggests that the features contain semantic information despite only being trained with RGB data.

We now apply this process to SRNs using a standard semi-supervised training framework.
Fig.~\ref{fig:pipeline_overview} summarizes the proposed semi-supervised approach.
In the first step, we pre-train the weights of the hypernetwork \hn, the latent embeddings $\mathbf{z}_i$ of the object instances in the training set, as well as the weights of the differentiable rendering purely for image reconstruction. This requires only posed RGB images as well as their extrinsic and intrinsic camera parameters.
Subsequently, we freeze $\mathbf{z}_i$ as well as the weights of \hn{} and the differentiable renderer, and train the proposed \emph{Segmentation Renderer} \seg{} on the learned feature vectors $\mathbf{v}$. This training is supervised with human-labeled, posed semantic segmentation masks of a small subset of the training images. 
In this case of limited training data, we parameterize \seg{} as a linear classifier in order to prevent overfitting. 

%% file: sections/analysis.tex
\begin{figure*}[t!]
	\centering
	\includegraphics[width=0.65\textwidth]{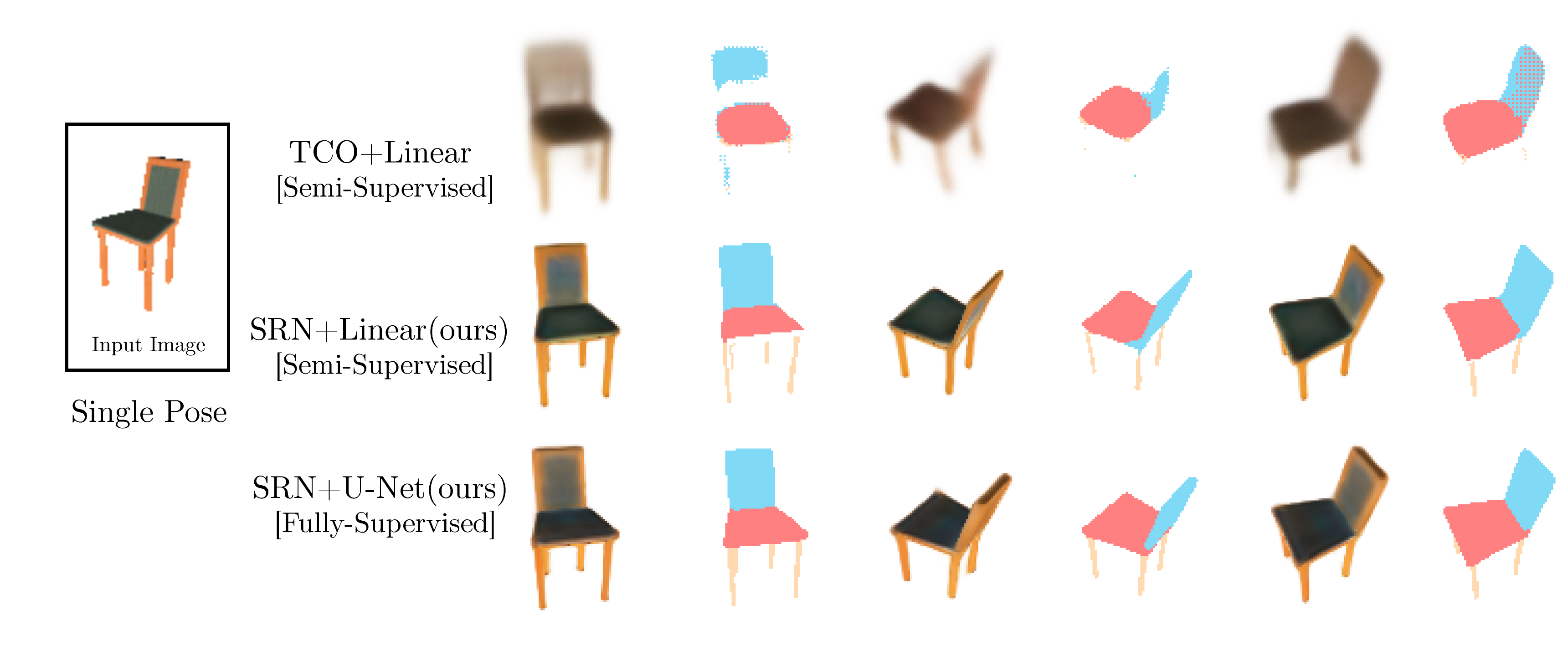}
	
	\caption{Comparison of all the \emph{single view} models, which can synthesize arbitrary RGB and segmentation views from only a single posed RGB image. The proposed semi-supervised \emph{SRN+Linear}  qualitatively outperforms the baseline semi-supervised approach by Tatarchenko et al.~\cite{TatarchenkoDB16} (TCO) and is comparable to  the fully-supervised \emph{SRN+U-Net} approach in terms of 3D consistency and semantic segmentation.}
	\label{fig:single_view}
\end{figure*}

\section{Analysis}
\label{sec:analysis}

In this section, we demonstrate that the proposed semi-supervised representation learning approach, using SRNs as the backbone 3D-structure-aware neural scene representations, succeeds in dense 3D semantic segmentation.
Model code and data are available at \url{https://www.computationalimaging.org/publications/semantic-srn/}. Specific implementation details including the model architectures, computational resources used, and training procedures can be found in the supplement.

Our ultimate goal is to learn a single, compact representation that jointly encodes information about 3D geometry, appearance, and semantic segmentation. 
To do so, we rely on comparisons in image space since, by design, this is the only data we have access to. We stress that this is merely a surrogate to demonstrate that the 3D representation contains semantic information, and not an attempt at incremental improvement on 2D semantic segmentation.
While it is possible to achieve similar input--output behavior with 2D approaches by building a pipeline that first leverages SRNs for novel view synthesis and subsequently feeds the image to a 2D model, this does \emph{not} demonstrate a multi-modal 3D representation. Instead, it encodes 3D information in the SRNs representation and semantic information in the 2D architecture. This does not support simultaneous reasoning about multiple modalities in 3D, which is critical to many real-world computer vision tasks (e.g., robotic grasping). We thus refrain from comparisons to such baselines.

%%
%In our first experiment, we train both fully- and semi-supervised instances of the proposed approach, demonstrating multi-view consistent RGB and semantic novel view synthesis. 
%%
%We benchmark these models with fully supervised and an alternative 3D representation learning approach.
%%
%Second, we demonstrate latent space interpolation, jointly interpolating semantic labels, texture and geometry.
%%
%Finally, we exploit that the proposed approach allows to infer a scene representation from both RGB and semantic mask modalities, and demonstrate reconstruction of geometry and texture from a single posed semantic segmentation image.
%%
%%
%
\begin{comment}
\par
\textbf{Implementation}
We implement all models in PyTorch. We train SRN-based models on Nvidia RTX8000 GPUs, and other models on Pascal TitanX GPUs.
%
The \srn{} as well as the \emph{RGB Renderer} are implemented as 4-layer MLPs with 256 units each, ReLU nonlinearities, and LayerNorm before each nonlinearity.
%
The raymarcher is implemented as an LSTM~\cite{Hochreiter:1997} with 256 units. We ray march for 10 steps. We train our models using ADAM with a learning rate of $4\mathrm{e}{-4}$. 
%
SRN-based models are trained for 20k steps at a resolution of 64 with a batch size of 92, and trained for another 85k steps at a resolution of 128 with a batch size of 16. Image reconstruction loss and cross-entropy loss are weighted $200:8$, such that their magnitudes are approximately equal.
\end{comment}
\par
\textbf{Dataset}
For all experiments, we use the PartNet~\cite{mo_partnet:_2018} and ShapeNet~\cite{chang_shapenet:_2015} datasets, which contain 3D meshes as well as their human-labeled semantic segmentation for a variety of object classes.
We conduct experiments using the chair and table classes with 4489 and 5660 object instances in the training set, 617 and 839 in the validation set, and 1214 and 1656 in the test set respectively.
Partnet contains labels at several resolutions. We conduct all experiments at the coarsest level of segmentation, leading to 6 chair and 11 table semantic classes.
We render observations using the Blender internal rasterizer.
For training and validation sets, we render 50 camera perspectives sampled at random on a sphere around each object instance.
For the test set, we render 251 camera perspectives sampled from a sphere around each object instance.

\par
\textbf{Evaluation.}
For quantitative evaluation of segmentation accuracy in image space, we adopt the mean pixel intersection over union (mIOU) and shape mIOU metrics used in \cite{mo_partnet:_2018}. 
For mIOU, we first calculate the average intersection over union across all the classes for each image seperately and then compute the mean of these mIOUs over all images and instances. 
In contrast, for shape mIOU, we first average intersection over union scores across all images and instances for each class separately and then average the class mIOUs. The shape mIOU score is generally much lower due to rare semantic classes which appear only in a small subset of all instances. These rare classes have small final class mIOUs and thus significantly reduce the average.

\begin{figure*}
	\includegraphics[width=\textwidth]{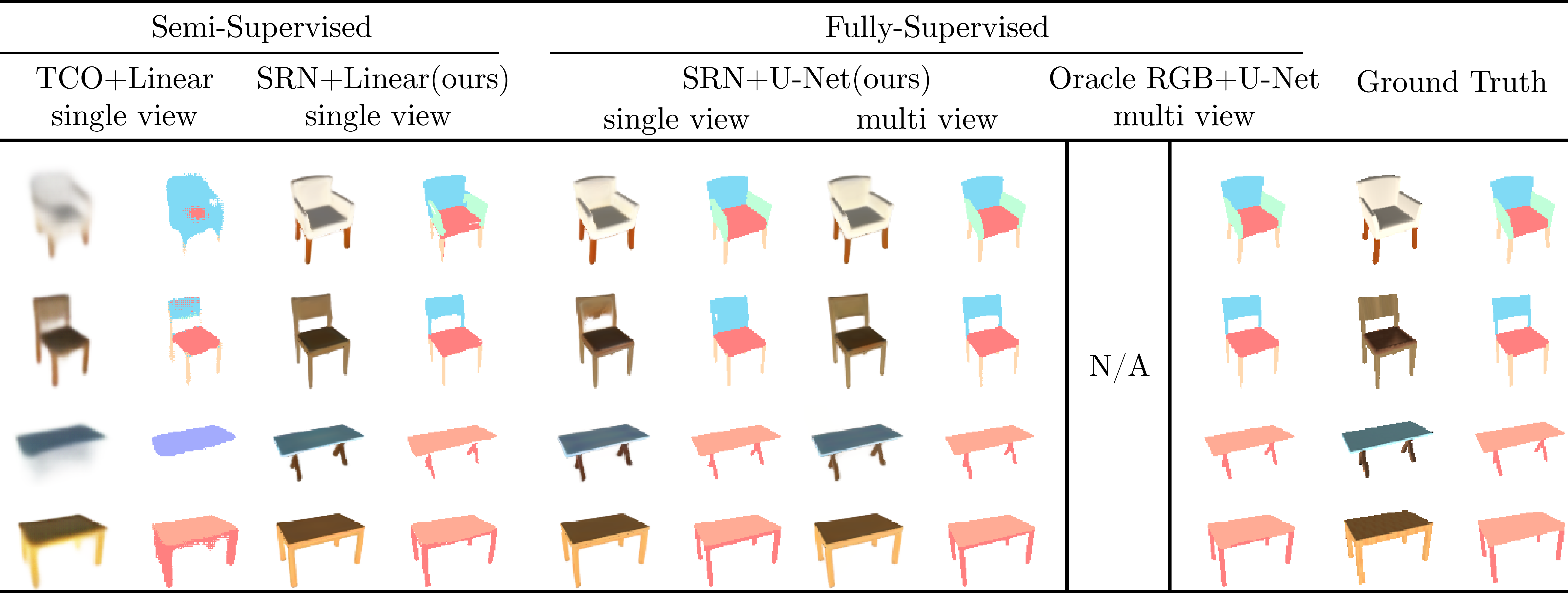}
	\caption{Qualitative comparisons of the approaches described in Sec. \ref{sec:analysis}.
		\emph{Semi-supervised approaches} (left column) include our proposed SRN+Linear as well as the baseline TCO+Linear model. After RGB only pre-training, these models are trained for segmentation with just 30 RGB images and their corresponding semantic maps. At test time, these methods receive a single posed RGB view. Please see the supplement for semi-supervised \emph{SRN+Linear} multi-shot results.
		\emph{Fully supervised approaches} (center column) include the SRN+U-Net reference model as well as the oracle RGB+U-Net. These models are trained on the full training corpus of RGB images and their per-pixel semantic maps. At test time, SRN+U-Net receives either a single or multiple RGB views while \emph{Oracle RGB+U-Net} receives all ground truth RGB views of the object.}
	\label{fig:qualitative_comparison}
\end{figure*}

\begin{table*}
	\begin{adjustbox}{width=\textwidth}
		\centering
		\begin{tabular}{@{}lccrcccrc@{}}
			\toprule
			\phantom{ab}
			& \multicolumn{2}{c}{Semi-supervised, small dataset} 
			& \phantom{}
			& \multicolumn{1}{c}{Supervised, small dataset}
			& \phantom{} 
			& \multicolumn{3}{c}{Supervised, full dataset} 
			\\
			\cmidrule{2-3} \cmidrule{5-5} \cmidrule{7-9}
			& TCO+linear & SRN+linear (ours) 
			&& Oracle RGB+U-Net
			&& \multicolumn{2}{c}{SRN+U-Net} & Oracle RGB+U-Net \\ 
			& single view & single view 
			&& multi view 
			&& single view & multi view & multi view \\ 
			\midrule
			Chairs & $28.4$ / $23.3$	& $48.7$ / $42.3$
			&& $42.2$ / $38.0$
			&& $60.9$ / $51.8$  & $74.2$ / $63.7$ & $77.3$ / $66.0$ \\
			Tables & $32.8$ / $11.4$	& $58.7$ / $18.3$
			&& $50.3$ / $17.9$
			&& $70.8$ / $26.5$  & $78.9$ / $40.5$ & $81.0$ / $44.7$ \\
			\bottomrule
		\end{tabular}
	\end{adjustbox}
	\vspace{1pt}
	\caption{Quantitative comparison of semi-supervised and supervised approaches. 
		We benchmark methods on mIOU as well as shape-mIOU.
		\emph{Semi-supervised} approaches (left column) as well as the \emph{Supervised, small-dataset} baseline are trained on 10 randomly sampled instances, 3 observations each. 
		\emph{Supervised, full dataset} (center column) baselines are trained on all training examples.
	}
	\label{tbl:one_shot}
\end{table*}

\subsection{Semi-supervised semantic segmentation.}
We experimentally evaluate the proposed multi-modal, 3D-aware neural scene representation (SRN+Linear) and compare it to related approaches. We demonstrate dense 3D semantic segmentation from extremely few labels, given only a single 2D observation of an object,  which allows for multi-view consistent rendering of semantic information. 
\par
\textbf{SRN+Linear.}
As discussed in Sec. \ref{subsec:semisup}, we first pre-train one scene representation network per object class to obtain a 3D-structure-aware neural scene representation.
We then pseudo-randomly sample 10 object instances from the training set such that all semantic classes are present.
For each of these instances, we randomly sample 3 posed images resulting in a total of 30 training examples. 
Following the proposed semi-supervised approach, we now freeze the weights of all neural networks and latent codes. We train a linear classifier to map features at the intersection points of camera rays with scene geometry to semantic labels.
\par
\textbf{TCO+Linear.} We benchmark the proposed method with a semi-supervised approach that uses Tatarchenko et al.~\cite{TatarchenkoDB16}, an auto-encoder-based neural scene representation, as the backbone.
We pre-train this architecture for novel-view synthesis on the full training set to convergence of the validation error, and then retrieve features before the last transpose convolutional layer.
We then train a single linear transpose convolutional layer on these features with the same 30 training examples used in the proposed SRN+Linear for direct comparison.
\par
\textbf{SRN+U-Net.}
As a 3D-structure aware reference model, we train the proposed model end-to-end with a U-Net segmentation classifier (see Sec.~\ref{sec:method}) on the full training dataset.
While this is not a semi-supervised approach, it yields an upper bound of segmentation accuracy of an SRN-based approach in a fully supervised regime of abundant labeled training data.
Note that this reference model does~\emph{not} infer a compact, multi-modal 3D-aware representation. Instead, this model may perform semantic segmentation in image space, and thus does not force the representation to encode all the information necessary for 3D semantic reasoning.
\par
\textbf{Performance.}
We first demonstrate that the proposed method (SRN+Linear) enables single-shot reconstruction of a representation that jointly encodes color, geometry, and semantic information.
Fig.~\ref{fig:single_view} shows the output of TCO+Linear baseline, the proposed semi-supervised SRN+Linear, and the end-to-end trained fully supervised reference model SRN+U-Net. 
SRN+Linear succeeds in multi-view consistent, dense 3D semantic segmentation. It far outperforms TCO+Linear and is comparable to the reference, SRN+U-Net.
In contrast, lacking a 3D-structure-aware representation, TCO+Linear fails to perform multi-view consistent semantic segmentation. 
The first four columns of Fig.~\ref{fig:qualitative_comparison} show further qualitative results for dense 3D semantic segmentation given \emph{single} and \emph{multiple} input views. 
Finally, Table~\ref{tbl:one_shot} shows quantitative results for the discussed methods. Consistent with qualitative results, the proposed SRN+Linear outperforms TCO+Linear and even approaches the performance of the single view, fully-supervised SRN+U-Net (see Table~\ref{tbl:one_shot}, column~4 and Fig.~\ref{fig:single_view}). While the proposed model's linear classifier sometimes struggles with parts of objects with higher inter-instance variance, it performs similarly to the reference models on common parts of objects, such as backrests, legs or the seat in the case of chairs. SRN+Linear operates in the most difficult regime of single view reconstruction with semi-supervision and still performs comparable to the SRN reference models.

\begin{figure*}
	\centering
	\includegraphics[width=0.75\textwidth]{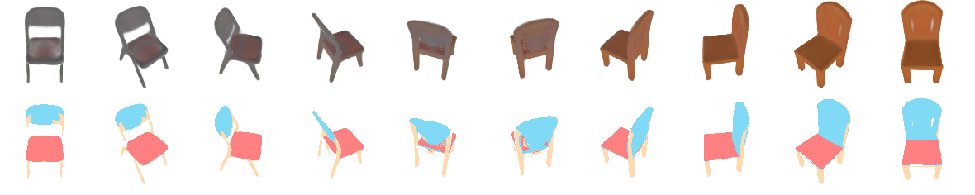}
	\caption{Interpolating latent code vectors while tracking the camera around the model. Both semantic labels and color features transition smoothly from object to object, demonstrating a tight coupling of semantic labels, geometry and texture of the objects.}
	\label{fig:interpolation}
\end{figure*}

\subsection{2D reference models with novel-view oracle.} 
As an upper bound for semantic segmentation performance, we consider the task of 2D-only semantic segmentation on the ground-truth RGB rendering of each test view.
\par
\textbf{Oracle RGB+U-Net.}
We first train a modern U-Net architecture on all pairs of images and their per-pixel semantic labels in the training dataset. At test time, we feed this architecture with a ground-truth RGB rendering of each test view.
%
\begin{comment}
Parameters of the U-Net are approximately matched with the proposed SRN-based approach.
%
\VS{Move U-Net architecture details to supplement}
Each downsampling layer consists of one stride-one convolutional layer, followed by one stride-two convolutional layer.
%
Each upsampling layer consists of one stride-two transpose convolutional layer, followed by one stride-one convolutional layer.
% 
We use BatchNorm and LeakyReLU activations after each convolutional block and dropout with a rate of $0.1$.
%
We train this model using the Adam optimizer with a learning rate of $4\mathrm{e}{-4}$ and a batch size of 64 until convergence of validation error after about 80k iterations or 20 epochs.
\end{comment}
We additionally train the reference 2D U-Net on the same 30 image-semantic-label pairs that the proposed semi-supervised approach is trained on.
In order to prevent the model from over-fitting, we use the validation set to perform a hyper-parameter search over dropout rates and use early-stopping. 
\par
\textbf{Performance.}
As expected, this oracle model trained on all the data (Table~\ref{tbl:one_shot}, column~6) outperforms the SRN reference models and the proposed semi-supervised method. However, it exists in the easiest regime of all the models, having access to the full dataset of segmentation maps for training and all the oracle RGB views at test time. 
Qualitatively, for more common objects in the test set, SRN+U-Net and the proposed SRN+Linear actually perform comparably to the oracle model, despite receiving only a small subset of the total information at both train and test time. 
Furthermore, the proposed models are able to perform the task of generating novel appearance and semantic segmentation views from a single observation, which the 2D-only oracle model cannot even evaluate as it does not support predictions about parts of the object that are occluded in the input view.
However, due to performing 3D reconstruction in addition to semantic segmentation, the proposed method fails whenever 3D reconstruction fails. This may be the case for out-of-distribution objects (see supplemental video). 
This failure mode is completely absent from the 2D oracle method as it does not need to perform any 3D reasoning.

The oracle model trained on the small 30 example training set (Tab.~1, column~3) is outperformed by the proposed semi-supervised method despite using additional segmentation data beyond the 30 training examples in order to perform early-stopping, and having access to the RGB novel-view oracle at test time.
This baseline does not have the compact, 3D multi-modal representation of the proposed method, and thus fails to generalize to other instances of the same class and maintain 3D-consistent views.

\begin{figure}
	\centering
	\includegraphics[width=.9\linewidth]{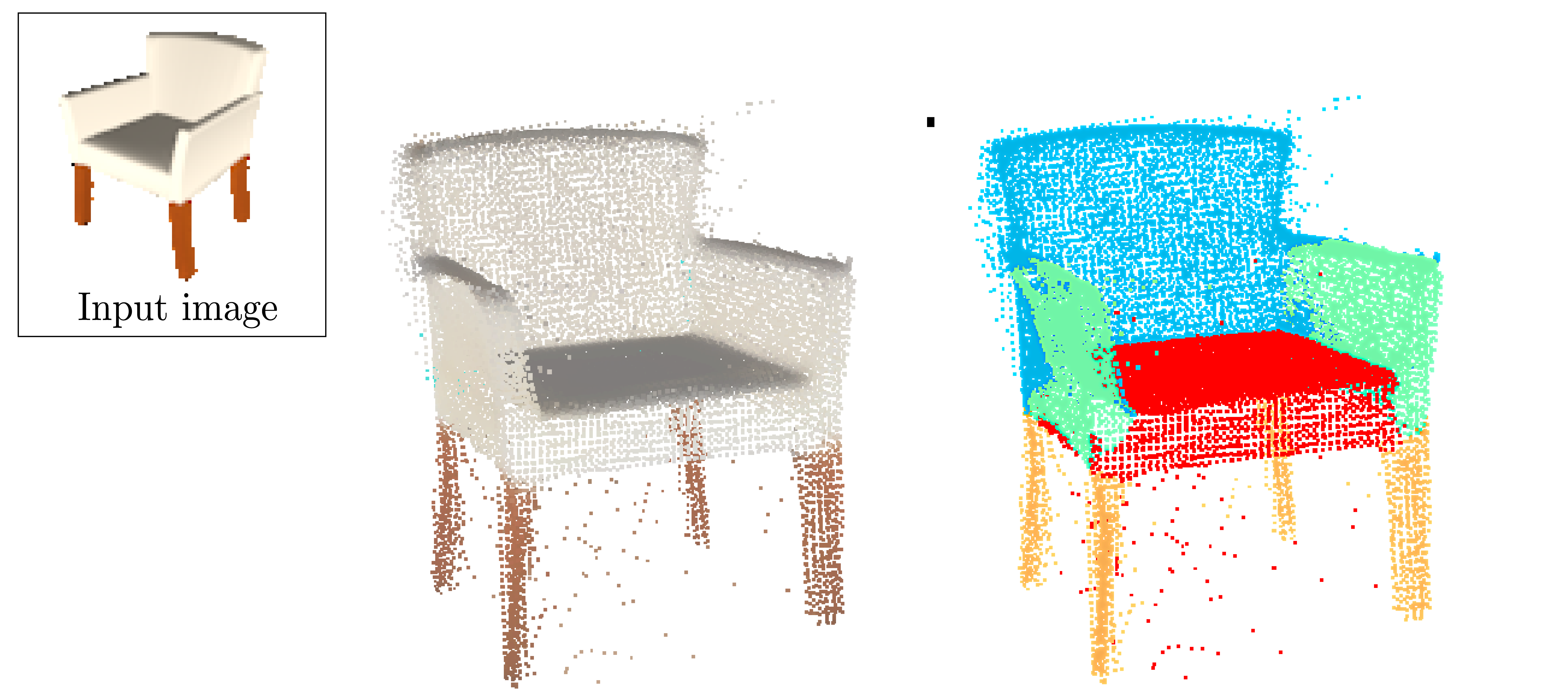}
	
	\caption{3D point clouds sampled from the continuous representation of an unseen chair from the test set, reconstructed by the proposed 30 training example SRN+Linear method from only a single input RGB image.}
	\label{fig:pc}
\end{figure}

\subsection{Additional Results.} 
In addition to dense 3D semantic segmentation, our proposed method can perform a number of additional tasks, which we subsequently explore.
\par
\textbf{Single training example.}
As a modification of our proposed SRN+Linear method, we train the segmentation linear classifier with only a single pose of a single chair instance (one image) instead of the usual 30 images. 
The model has a 2.7\% increase in mIOU and a 4.2\% decrease on shape mIOU compared to our original 30 training example method on the full test set. Moreover, the model correctly labels the back legs and backs of chairs in the test set, despite having never seen them at train time. 
The quantitative result was expected since our 30 example model includes rare classes whereas our single example has only common class labels. This means that the single example model does better for common chairs (higher mIOU), but fails to classify the unobserved classes (lower shape mIOU). Qualitative results can be found in the supplemental video.
\par
\textbf{Instance Interpolation.}
Interpolating latent vectors inferred in the proposed framework amounts to jointly interpolating geometry, appearance and semantic information. 
Fig.~\ref{fig:interpolation} visualizes a latent-space interpolation of two chairs in the test set, both reconstructed from a single view by the proposed semi-supervised linear model. 
Geometry, appearance and semantic labels interpolate smoothly, demonstrating a tight coupling of these modalities.
\par
\begin{figure}
	\centering
	\includegraphics[width= \linewidth]{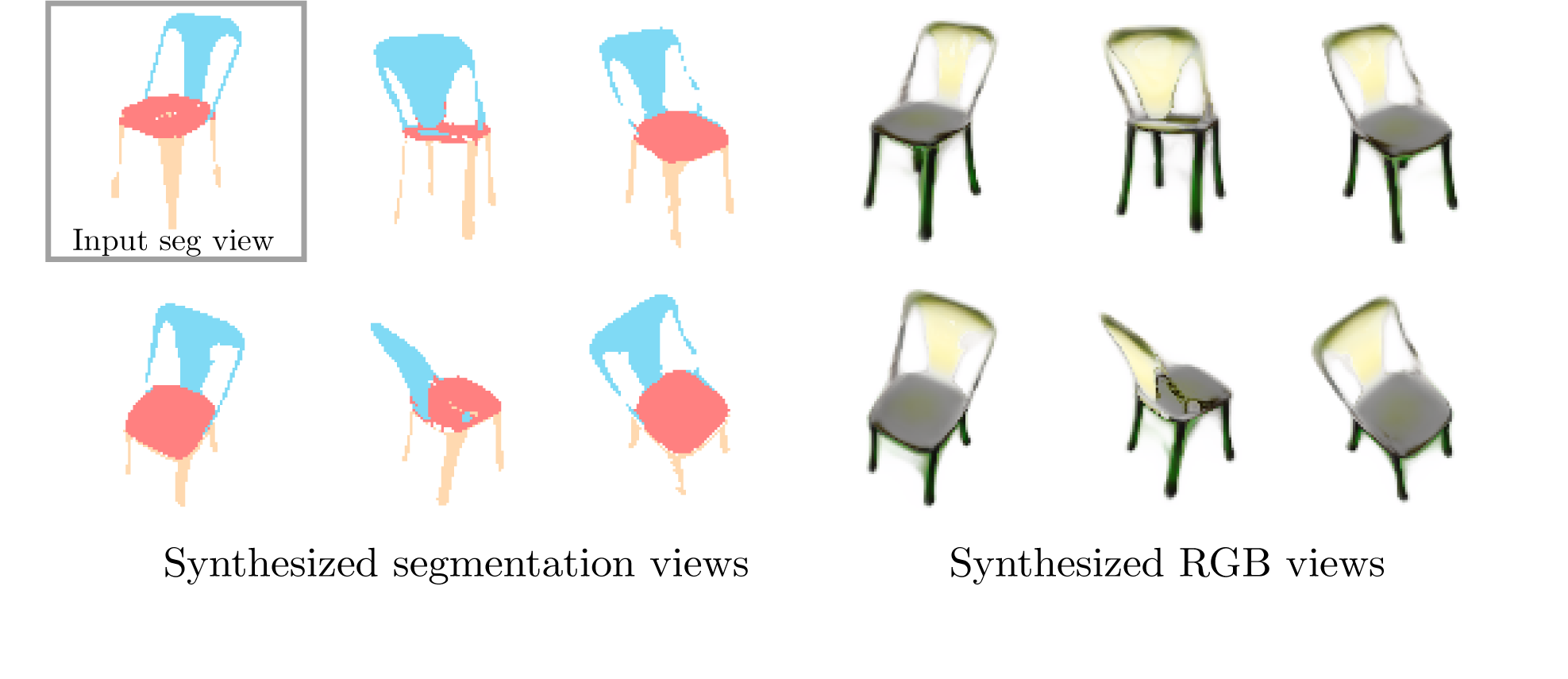}
	\caption{The proposed methods may infer a neural scene representation either from RGB images or semantic segmentation masks, or both. Here we show renderings reconstructed from a single segmentation mask, using the proposed SRN+Linear. 
	}
	\label{fig:seg2rgb}
\end{figure}

\textbf{3D reconstruction from semantic mask.}
As an instantiation of the auto-decoder framework~\cite{ParkFSNL2019}, inferring the neural scene representation of a novel object amounts to initializing and subsequently optimizing a new embedding vector to minimize reconstruction error.
As all the proposed methods may be supervised by both semantic segmentation labels and RGB renderings, they also enable reconstruction of neural scene representations through either modality. 
Fig.~\ref{fig:seg2rgb} visualizes 3D reconstruction of a chair from a single posed segmentation mask, while Fig.~\ref{fig:teaser} demonstrates reconstruction from a single posed color image.
\par
\textbf{3D Point Cloud rendering.}
In addition to multi-view images, our proposed SRN+Linear method is also able to produce point clouds of arbitrary resolution.
The process is nearly identical to the proposed method. The only difference is that we collect the points given by the ray marcher across multiple poses instead of forming an image per each pose. We then query the SRN at each point in this set to obtain both RGB and semantic label point clouds. Fig. \ref{fig:pc} shows the result when we sample 65,165 points for an unseen chair in the test set.

\begin{comment}
\begin{figure}
	\centering
	\includegraphics[width=\linewidth]{figures/failure_cases_new_v1.pdf}
	\caption{Failure cases.}
	\label{fig:failure_cases}
\end{figure}

\subsection{Failure cases.}
Fig.~\ref{fig:failure_cases} displays failure cases of the proposed approach.
%
The proposed approach inherits limitations and failure cases of scene representation networks, such as failure to reconstruct strong out-of-distribution samples or objects with small gaps or high-frequency geometric detail. 
%
In these cases, the semantic segmentation fails as well.
%
In the semi-supervised regime, the linear classifier sometimes fails to assign the correct class even if geometry and appearance were reconstructed correctly, which we attribute to its limited representative power. We note that as both appearance-based 3D neural scene representation methods as well as semi-supervised representation learning methods further develop, these failure cases will improve.
\end{comment}

%% file: sections/discussion.tex
We present a 3D representation learning approach to joint reconstruction of appearance, geometry, and semantic labels. 
Our semi-supervised method only requires 30 human-annotated, posed semantic segmentation masks for training.
At test time, this enables full 3D reconstruction and dense semantic segmentation from either posed RGB images, semantic segmentation masks, or both, from as few as a single observation. 

Our method contains failure cases including out-of-distribution objects, instances with rare classes, and cases where the SRN fails to reconstruct the scene. A  detailed qualitative overview of these failure cases and the limitations of our methods can be found in the supplement and supplemental video.

We believe that our work outlines an exciting direction in extending both scene representations and representation learning methods. As both of these fields independently develop more powerful techniques, we expect that our proposed technique will also improve.

Future work may extend the proposed 3D-aware representation learning approach to generalize other scene properties, such as affordance, material properties, mechanical properties, etc. across a class of scenes given extremely few observations. We also hypothesize that the proposed approach will generalize to room-scale environments, where it would enable scene semantic segmentation given extremely few labels.